\documentclass[11pt]{article}
\usepackage{acl2016}
\usepackage{times}
\usepackage{url}
\usepackage{latexsym}
\usepackage{graphicx}
\usepackage{float}

\aclfinalcopy % Uncomment this line for the final submission
\title{Content Selection in Data-to-Text Systems: A Survey}

\author{Dimitra Gkatzia \\
  Edinburgh Napier University \\
  {\tt d.gkatzia@napier.ac.uk} }

\date{}

\begin{document}
\maketitle
\begin{abstract}

Data-to-text systems are powerful in generating reports from data automatically and thus they simplify the presentation of complex data. Rather than presenting data using visualisation techniques, data-to-text systems use natural (human) language, which is the most common way for human-human communication. In addition, data-to-text systems can adapt their output content to users' preferences, background or interests and therefore they can be pleasant for users to interact with. Content selection is an important part of every data-to-text system, because it is the module that determines which from the available information should be conveyed to the user. This survey initially introduces the field of data-to-text generation, describes the general data-to-text system architecture and then it reviews the state-of-the-art content selection methods. Finally, it provides recommendations for choosing an approach and discusses opportunities for future research.
\end{abstract}

\section{Introduction}

\begin{figure}
\center
 \includegraphics [width = 8cm]{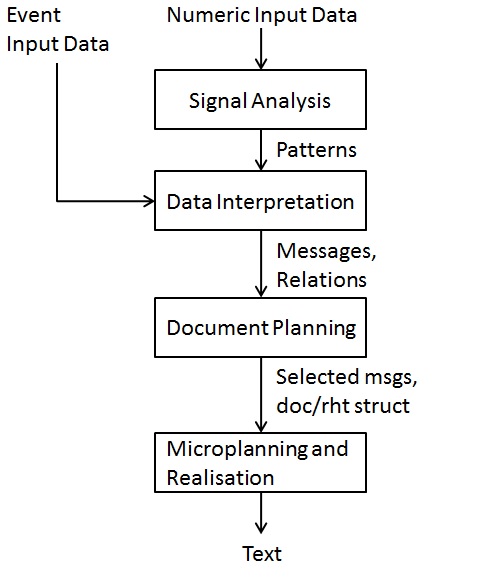}
 \caption{Data-to-text system architecture (Reiter, 2007) .}
 \label{d2t}
\end{figure}

\textit{Natural Language Generation (NLG)} is the sub-area of Natural Language Processing (NLP) which focuses on building software systems that generate text or speech in human languages. Typically the input of an NLG system is some non-linguistic representation, such as knowledge bases or numerical data. \textit{Data-to-text} generation is the area of Natural Language Generation (NLG) which elaborates the summarisation of numerical data, such as time-series data from sensors or event logs \cite{Reiter2000}. This survey particularly focuses on \textit{time-series data}, i.e. data which change over time. Examples of such data are sensor data (e.g.\ heart rate, breathing rate), weather data, stock market data, etc. which can be found in modern applications on mobile phones and wearable gadgets as well as traditional desktop applications. Reiter \shortcite{Reiter2007} proposes a general architecture for data-to-text systems which consists of four distinct modules: (1) Signal Analysis, (2) Data Interpretation, (3) Document Planning and (4) Microplanning and Realisation. The four modules are described briefly below and their relations are depicted in Figure \ref{d2t}:
\begin{enumerate}
\item{\textbf{Signal Analysis:}} The Signal Analysis module is responsible for analysing the input data, identifying patterns and trends. This is an essential part of a data-to-text system when the input is numerical data.
\item{\textbf{Data Interpretation:}} The Data Interpretation module is responsible for detecting causal and other relations between the patterns and trends identified by the Signal Analysis module. This module is useful for NLG systems that aim to communicate more complex messages, such as explanations.
\item{\textbf{Document Planning:}} This module decides which of the identified patterns, trends and relations should be conveyed in the generated textual summary, a task known as \textit{\textbf{content selection}}. It is also responsible for structuring the generated text, i.e. deciding on information ordering, the paragraph breaks in longer generated documents and the general structure of a document. The document planner is essential when part of the available content needs to be communicated, such as in report generation or summarisation of time-series data.
\item{\textbf{Microplanning and Realisation:}} This module actually generates the output text. Every NLG system contains a realisation module, which can be either template based, i.e.\ canned text, or it can use sophisticated methods based on syntax, sentence structure and morphology.
\end{enumerate}

The focus of the survey is on the task of content selection from time-series data. In this survey, assumed applications generate summaries of a paragraph long or documents whose structure is predefined. 

Time-series data such as sensor data, weather data, stock market data etc., often has a complex internal structure and its analysis in order to derive the underlying information is domain-dependent. Humans can summarise time-series data effectively by using natural language. According to \cite{Gkatzia2014acl}, the descriptions of time-series data can be a result of an interplay of decisions, for example, one can decide to mention the unusual fluctuations, the averages or the time-series changes (trends) over time. Moreover, they can decide to refer to the data in a sensible order, or to use their broad knowledge to justify and/or explain the time-series data. In contrast, the task of automatically determining the content for effective summarisation of time-series data continues to be challenging, due to the fact that content determination in general is domain-dependent. The necessity of general content selection models has been acknowledged by the NLG community \cite{BouayadAgha2012}. 

\subsection{Why a survey on content selection, and how to read it?} Content selection approaches have been developed considerably since the overview presented by Reiter and Dale \shortcite{Reiter2000}. These new approaches have been presented and published in very different venues, which makes it hard for a newcomer to investigate all the available literature. In this single survey, there is work presented from over 30 different publication venues, ranging from computational linguistics to data mining, artificial intelligence and medical informatics. 

The added value of this survey is that it collects all relevant literature and discusses it in an informative way. In addition, directions for future work are suggested. Consequently, this survey is addressed to all researchers and practitioners of natural language generation, who aim to start working on or they resumed working on content selection. As it includes an informative introduction, it can be useful for students and researchers who are new to this field, as well as experienced researchers who want to catch up with latest developments. 

The next section describes the motivation of dealing with the task of content selection from time-series data (Section \ref{motivation}). Section \ref{challenges} presents the challenges of content selection in data-to-text systems. Then, the survey describes the related work in this area. It classifies the related work into two main areas: rule-based approaches \ref{ruleBased} and trainable methods \ref{statistical}. Section \ref{adaptiveSystems} discusses adaptive NLG systems, followed by a discussion on evaluation metrics (Section \ref{metrics}). Finally,  Section \ref{litConclusions} concludes the work presented in this survey and Section \ref{futureWork} presents opportunities for future research. Appendix A includes a list of the available datasets for data-to-text generation.

\section{Motivation} \label{motivation}
This section highlights the importance of content selection in data-to-text systems. In particular, it aims to answer: (1) why it is important for data-to-text systems to determine the relevant content (Section \ref{simplify});  (2) why it is preferable to present textual summaries of time-series data rather than graphical representations  (Section \ref{vsgraphical}); and (3) why it is important for content selection to be user-adaptive (Section \ref{adapt}).  

\subsection{Achieving the Communicative Goal}\label{simplify}
Data-to-text systems are useful for simplifying the presentation of complex data. The overload of time-series data available through the web, sensors and other means has increased the need of digesting these data in an automatic, accurate and time-efficient manner. NLG systems can deal with this challenge in an automatic and fast way. For example, an NLG system can read sensor data and produce a comprehensive textual summary.  Previous research on generation from time-series data has been conducted in several domains such as weather forecasts \cite{Sripada2004,Konstas2012,Gkatzia2016}, health informatics \cite{Gatt2009,Gkatzia2016fuzz}, stock market summaries \cite{Kukich1983} and assistive technology systems \cite{Black2010}. These systems have employed different content selection methods, which are reviewed in Section \ref{ruleBased} and \ref{statistical}. 
In addition, it is essential for data-to-text systems to choose the relevant information to communicate so as to satisfy the users' preferences or to assist in decision making \cite{Gkatzia2016}.

\subsection{Effectiveness of Textual Summaries over Graphical Representations}\label{vsgraphical}

Data and in particular time-series data is normally presented using visualisation techniques that can be difficult for an inexperienced user to understand. Data-to-text systems face the challenge of communicating the data in a simpler, more effective and more understandable way, by conveying information through language. Recent studies have showed that text descriptions of data can be more effective, comprehensible and helpful in decision making than the corresponding graphical representations  \cite{Meulen2010,Law2005}, even when the data is uncertain \cite{Gkatzia2016}.

Early research has shown that graphs require expertise in order to be interpreted \cite{Petre1995}. More recently, Law et al. \shortcite{Law2005} compare expert-generated text summaries with the corresponding pattern graphs in relation to their effectiveness to support clinical decisions. In that study, clinical staff are shown text summaries and graphical representations of patient's data and are asked to make a decision about the health condition of the patient. It is found that clinical staff tend to make correct clinical decisions when viewing the textual format of the data rather than when consulting the graphs. 

van den Meulen et al. \shortcite{Meulen2010} describe a similar survey which focuses on the effectiveness of text compared to graphs in decision making in the health domain. The evaluation process follows the one described by \cite{Law2005}. This study shows that, although users prefer the textual descriptions produced by humans, the computerised textual reports are as useful in decision making as the graphs that the staff were familiar with.

Hunter et al.\ \shortcite{Hunter2011} focus on the potential of improving clinical decisions by employing NLG. Indeed, their approach is evaluated by presenting nurses a computer generated summary with the corresponding data graph. Then, nurses are asked to make a clinical decision. It is found that exposure to textual summaries lead to comparable decision making as with the graphical representations. Gatt et al.\ \shortcite{Gatt2009} present a system evaluation in the same domain. It is shown that all the users (doctors and nurses) perform better in parallel tasks (making decisions after viewing a text summary vs.\ a graphical representation) with human-written texts rather than graphs. Compared to computer-generated summaries, they perform worse than they perform with the hand-written texts, however the users find the computer-based texts as useful as the graphs. 

More recently, Gkatzia et al.\ \shortcite{Gkatzia2015enlgdemo} presented a game whose aim is to measure decision making under uncertainty with different conditions (graphical representations vs.\ text vs.\ multi-modal representation). It was found that for specific user groups text descriptions of weather data led to better decision making than other descriptions \cite{Gkatzia2016}.

These examples of previous research show that the interpretation of graphical representations is not always obvious. It also shows that textual descriptions can effectively support and enhance decision making. Although we acknowledge the importance of combining text and visualisations (e.g. \cite{Gkatzia2016,Mahamood2014,Sripada2007}), in this survey we focus only on automatic content selection from time-series data.

\subsection{Adaptive Output}\label{adapt}

Different user groups such as doctors, nurses and parents or lecturers and students have different information needs and preferences, therefore personalised reports are important. Hunter et al.\ \shortcite{Hunter2011} emphasise that personalisation should be based on relevant factors/ variables as for instance users' stress level, and not only demographic data, although in some cases  demographic data can be important to adaptation. Users have different preferences and goals and the systems should adapt to those in order to be more preferable. DiMarco et al.\ \shortcite{DiMarco2008} emphasise that it is necessary for a system to avoid referring to events that seem irrelevant for the majority of the users, but are relevant for a particular user. For example, today's health care systems can provide too much irrelevant information to patients or omit important information, which leads the users to believe that the system is not addressed to them \cite{DiMarco2008}. This can have, for instance, negative impact on the patients' compliance with medical regimens. Similarly, in the student feedback domain, a general system would advice students to study $x$ hours per day. For a hard-working student this advice might be irrelevant or even disturbing \cite{GkatziaThesis}.

On a related note, Gkatzia et al. \shortcite{Gkatzia2013enlg} show that there is a mismatch between the preferences of students and lecturers on what constitutes a good feedback summary and therefore they introduce a new task, Multi-adaptive Natural Language Generation (MaNLG), which aims to find a balance between the preferences (or other conditions of interest) of two different user groups, as for instance, lecturers and students, or patients and doctors \cite{Gkatzia2014inlg}. 

\section{Challenges for Content Selection in Data-to-text Systems} \label{challenges}
In this section, the challenges for content selection in data-to-text systems and recommendations for addressing them are presented.

\begin{itemize}
\item{\bf{Data availability:}}
The lack of aligned datasets (data and corresponding summaries) that can be used to derive rules or to train an NLG system is a major challenge for NLG engineers. Although data are widely available, they cannot be used directly for the development of an NLG system, because there is lack of alignment between input and output data \cite{BelzKow2010}. Data-driven data-to-text systems require large corpora with data that can be aligned to natural language text so as to be used as an input to a training algorithm. 
In Appendix \ref{datasets}, we offer a list of freely available datasets for data-to-text generation.

\item{\bf{Domain dependence:}}
Data-to-text systems are domain sensitive which makes it hard to transfer modules or data between domains.

\item{\bf{Evaluation challenges:}}
As other areas of Computational Linguistics, NLG also suffers from the limitations of the available evaluation methods. Reiter and Sirpada \shortcite{Reiter2002} firstly questioned the suitability of corpus-based approaches to evaluation of NLG systems, followed by \cite{Foster2006} and \cite{Belz2006}. Text corpora from data are usually gathered by asking experts to provide written textual summaries or descriptions. However, experts use different words to communicate the data or they choose to refer to different events, which makes it difficult to construct a consistent dataset and therefore using it as gold standard for evaluation. 

\item{\bf{Lack of or inconsistent expert knowledge:}}
Another issue is the lack of expert knowledge or the difficulties of acquiring it due to several factors, such as difficulties in recruiting experts. The main challenge is that experts provide a variety of responses, which introduces difficulties in knowledge acquisition. This challenge has been also noted by \cite{Sripada2004}.  

\item{\bf{Lack of prior knowledge of the users:}}
One of the most crucial issues in adaptation is the lack of prior knowledge of the users. This issue has been raised by several researchers, such as \cite{SriniThesis,Han2014}, to name a few. Previous approaches to tackling this issue include the use of latent User Models \cite{Han2014}, initial questionnaires to derive information by the user \cite{Reiter1999} and tackling first-time users using multi-objective optimisation \cite{Gkatzia2016fuzz}. 
\end{itemize}

Although these are important challenges, there are ways to address them. The data availability issue can only be solved by creating parallel corpora for every new domain, or investigate approaches which allow the transfer of knowledge from one domain to another. The evaluation of data-to-text systems can be improved by performing human evaluations, and in particular task-based evaluations. The lack of prior knowledge of the user can be solved by investigating approaches which aim at addressing to unknown users.

\section{Rule-based Content Selection in Data-to-Text Systems} \label{ruleBased}

This section describes previous work that treats content selection in a rule-based manner. Table \ref{rulebasedTable} summarises the methods and application domains of these systems. One of the earliest data-to-text applications is {\tt TREND} \cite{Boyd1998}. {\tt TREND} includes a very detailed module for time-series analysis using wavelets. However, this system does not include a notion of content selection, as it is mostly focused on describing all trends that are observed in data.

\begin{table*}[ht!]
\centering
\caption{Content selection in rule-based data-to-text systems\label{rulebasedTable}}
\begin{tabular}{|p{4.8cm}| p{3.7cm} |p{2.6cm}|p{2.9cm}|} % centered columns (4 columns)
\hline %inserts double horizontal lines
Author(s) & Method & Domain & Data Source \\  % inserts table 
%heading
\hline \hline% inserts single horizontal line
\cite{Boyd1998} & No content selection  & Weather & database\\ \hline
\cite{Sripada2001} & Two stage model:  (1) Domain Reasoner and (2) Communication Reasoner& Weather, Oil rigs & sensors, numerical data\\ \hline
\cite{Sripada2003} & Gricean Maxims & Weather, Gas turbines, Health  & sensors\\ \hline
\cite{Hallett2006} & Rule-based & Health  & database\\ \hline
\cite{Yu2007} & Rules derived from corpus analysis and domain knowledge & Gas Tourbines & sensors\\ \hline
\cite{Sripada2007} & Decompression Models & Dive & sensors\\ \hline
\cite{Turner2008inlg} & Decision Trees & Georeferenced Data & database \\ \hline
\cite{Gatt2009} & Rule-based & Health & sensors\\ \hline
\cite{Thomas2010} & Document Schemas & Georeferenced Data &database\\ \hline
\cite{Demir2011} & Rule-based & Domain independent & graphs - database \\ \hline
\cite{Peddington2011} and \cite{Tintarev2016} & Threshold-based rules & Assistive Technology & sensors\\ \hline
\cite{Johnson2011} & Search Algorithms &Autonomous Underwater Vehicle & sensors\\ \hline
\cite{Banaee2013} & Rule-based & Health & grid of sensors \\ \hline
\cite{Schneider2013} & Rule-based & Health & sensors\\ \hline
\cite{Soto2015}& fuzzy Sets & Weather & database \\ \hline
\cite{Gkatzia2016} & Rule-based & Weather & numerical data with assigned probabilities\\ \hline
\end{tabular}
\end{table*}

Sripada et al.\ \shortcite{Sripada2001} suggest a ``two-stage model for content selection'' from time series data (sensor readings from a gas turbine and numerical weather simulations). The model assumes that the data source is an external component and that a \textit{Domain Reasoner (DR)} module is evident. The DR is responsible for making inferences. The inferences together with the system's communicative goal are used for building an overview of the summary. Finally, the \textit{ Communication Reasoner} module takes as input the output of the DR and it specifies the final content, which is then available to the other NLG tasks, i.e. microplanning and surface realisation. It is worth mentioning that this approach is suggested before 2007, when a general data-to-text approach is introduced by \cite{Reiter2007}, as we discussed in the introduction.

Sripada et al.\ \shortcite{Sripada2003} introduces a domain-independent approach to Natural Language Generation using Gricean Maxims from the same data sources as previously mentioned plus medical sensor data. The Gricean maxims are used in order to communicate the content, after the segmentation algorithms have been applied for data analysis \cite{Sripada2003}. The maxims reflect the cooperative principle that describes how people communicate and act with one another, by using utterances, their flow and their meaning. The Gricean maxims \cite{Grice1975} constitute the Quality, Quantity, Relation and Manner maxims and they are inspired by the pragmatics of natural language. The maxim of Quality influences the content selection decisions regarding the real values of the data by using linear interpolation. The maxim of Quantity decides on which data patterns are useful for the user. The maxim of Relevance estimates which information might be relevant to a particular user and User Models are acquired for this task, as in \cite{Reiter2003}. Finally, the maxim of Manner influences the linguistic decisions, i.e.\ how the information should be conveyed, without ambiguity, briefness and in a sensible order.

Hallet et al.\ \shortcite{Hallett2006} present a content selection approach for summarisation of medical histories which is based on two elements: (1) the type of the summary and (2) the length of the summary. They also introduce a list of concepts and, events which are linked to those concepts. During the content selection phase, the events are clustered in terms of relevance. It is assumed that smaller clusters do not include important events and therefore, only the larger clusters of events are mentioned in the summary. Depending on the type and the length of the summary the content attributes are determined in a rule-based fashion. For instance, a problem might be a main event and its attributes can be name, status, clinical course etc.

Sripada and Gao \shortcite{Sripada2007} report the ScubaText system which generates reports from scuba-dive computer data and it detects the safety of the dives. The data analysis module determines the interpretations of the patterns identified regarding the safety of dives. Decompression models (similar to those used by dive computers) are used to generate recommendations on when the bottom is safe for diving. Using these interpretations, deviations from the actual dive are computed. Then, ratings are assigned inversely proportional to the deviations and they influence the text generation decisions.

Yu et al.\ \shortcite{Yu2007} present SumTime-Turbine, a system that summarises large time-series data sets from gas turbines' sensors. This system adopts a bottom-up approach, where the NLG system emerges by joining subsystems together. It consists of two main components: a data analysis module that is responsible for content selection  and the Natural Language Generation module. The data analysis component can be further split up into:
\begin{itemize}
\item Pattern Recognition, which is responsible for connecting time-series segments to concepts.
\item Pattern Abstraction, which maps patterns to abstract concepts .
\item Interesting Pattern Selection, which is responsible for deciding which of the abstract patterns should be conveyed in the summary. The content is determined by using domain knowledge and historical pattern frequency.
\end{itemize}
The content order is based on rules obtained via corpus analysis and is inspired by the way that experts tend to summarise sensor data. In particular, the content follows the following ordering.
\begin{itemize}
\item Background information
\item Overall Description
\item Most significant patterns
\end{itemize}

\begin{figure*}[!ht]
  \centering
      \includegraphics[width=0.8 \textwidth]{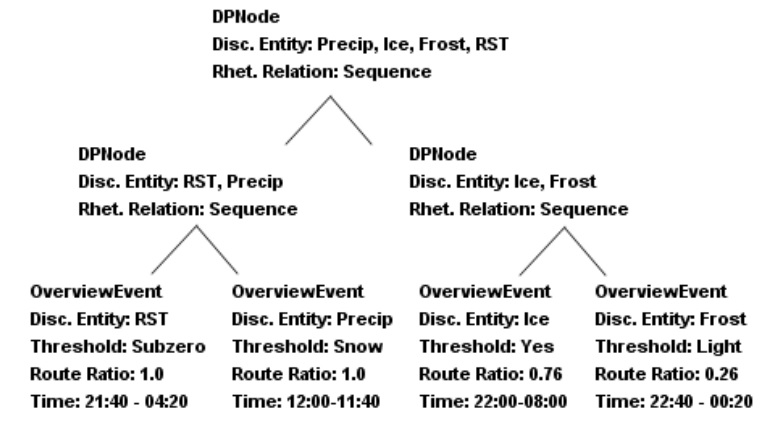}
  \caption{\label{RoadSafe} RoadSafe: Example of event tree \cite{Turner2008inlg}.}
\end{figure*}

Turner et al.\ \shortcite{Turner2008inlg} present a decision tree approach to content selection in the domain of description generation of geo-referenced data. In this framework, content is represented as leaves of a tree, whereas the nodes represent events. The text is then generated from the content that exists in leaves. Figure \ref{RoadSafe} shows the overview of events and content.  In a similar domain, \cite{Thomas2010} use document schemas to induce document plans for textual descriptions of geo-referenced data for blind users. The selection of the schema is influenced by the spatial data analysis.

% \begin{figure}[hH]
%\centering
% \includegraphics [width = 12cm]{RoadSafe.jpg}
% \caption{RoadSafe: Example of event tree \cite{Turner2008inlg}.}
% \label{RoadSafe}
%\end{figure}

Hallet et al.'s\ \shortcite{Hallett2006} algorithm is also used for content planning at the BabyTalk system \cite{Gatt2009}, which produces text summaries of neonatal intensive care unit data. The data used as input consist of (1) sensor data (Heart Rate, mean Blood Pressure and Oxygen Saturation), (2) lab results and observations, (3) events such as nurses actions, medical diagnosis and treatment and other information, and (4) free text. The BT-45 system generates a summary after 45 minutes of measurements and collection of the data mentioned earlier \cite{Portet2007}. Its aim is to interpret the data by linking events to observations, not to offer diagnosis. 

In this system, content selection is handled as described in \cite{Hallett2006}, where the length of the goal summary is a deciding factor as well as the type of problem.  The events are assigned an importance value which represents their significance in the data interpretation module, but it ignores the possibility of a value change after content selection, e.g.\ an event that explains a fact may be omitted, thus making the summary non-coherent. The events are clustered in terms of relevance and the first step of the summarisation dictates the removal of the smaller clusters, because they are usually irrelevant. Next, the important events and the level of details are influenced by the relevance to the type of summary. However, the system lacks in updating the importance of the events (regarding the probability to be selected) after one event is being selected. 

\begin{figure}[hH]
\center
 \includegraphics [width = 7.5cm]{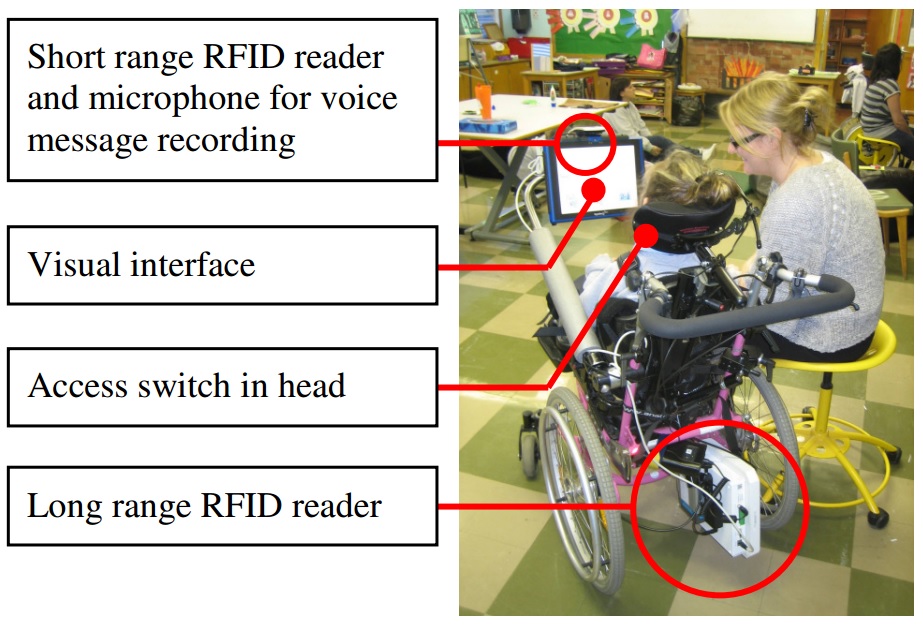}
 %\caption{The ``How was the school today...?'' system \cite{Black2010}.}
 \label{howWasSchool}
\end{figure}

Black et al.\ \shortcite{Black2010} develop a story generation rule-based system that is addressed to children with Complex Communication Needs. The input of this system is non-linguistic data gathered through sensors which describe the child's location, activities and interactions with people or objects. Specifically, the data are collected through: (1) RDIF readers which monitor the places that the child visits, (2) a microphone that is used for recording events, and (3) a visual interface and an access switch that the child can use with its head. The teacher and the school staff can also enter information about the child's activities. Figure \ref{howWasSchool} presents the overall structure of the system.

The goal of the system is to automatically generate a narrative about ``how was school today...''. The system groups elements into events in order to determine the content \cite{Peddington2011},  by using clustering algorithms to classify events depending on the location, the time and the voice recordings. It also employs rules to define unexpected events, for example the divergence with the child's usual timetable and activities. The derived rules are based on a User Model that takes into account the child's cognitive model, the timetable, unexpected events and inherent ``interestingness'' five events are selected out of twelve \cite{Tintarev2016}. 

Demir et al.\ \shortcite{Demir2011} present an approach to summarisation of bar charts. This is a domain-independent approach which is based on users' scores of potential content to be present in a brief summary. After averaging scores from a data collection, derived rules determine what information of a graph should be included in the summary.

Johnson and Lane \shortcite{Johnson2011} present Glaykos, a system that automatically generates audio visual debriefs for underwater missions. The data used are collected through an Autonomous Underwater Vehicle (AUV) that is armed with sensors. The sensor data describe attributes of the bottom of the sea. In order to generate the multimodal output, a situation model is used, which consists of a bitmap situation model and a vector situation model in order to include all the data from the underwater mission and the related simple and complex concepts. Initially, the data from the mission are processed and linked with the other elements of this bitmap situation model. From this, a vector situation model is created, which models the motivation and the causation models. Next, the events are ordered and grouped together according to the time they happened (adjacent time), whether they have the same actor, whether they do not have contradictory motivations and depending on their causal relations. These groups are viewed as an instance of the travelling salesman problem, where each group represents one city. In order to solve this, two optimisation algorithms have been applied, a depth-first search and a genetic algorithm. Both algorithms used the same fitness function, which is based on the spatiality of the event (a penalty is given if it is in a different region), temporality of the action, the protagonist, the motivation and the causality. 

Banaee et al.\ \shortcite{Banaee2013} present a content selection approach for summarisation of physiological sensor data based on the importance of potential content messages: (1) messages conveying holistic information, (2) messages conveying events, and  (3) messages which summarise events. Each message category uses a ranking function to assign an ``importance'' value to the message. The ordering of the message is based on how important the message is and whether there are dependencies between messages. 

Schneider et al.\ \shortcite{Schneider2013} describe an approach to summarising medical sensor data in pre-hospital care (MIME project). The content selection module is rule-based and it uses trees that associate the chosen information, inspired by the Rhetorical Structure Theory  \cite{Mann1988}. The rules are derived through a combination of corpus analysis and expert consultation. 

Soto et al. \shortcite{Soto2015} describes an approach to short weather forecasts generation using fuzzy sets. In their approach, content selection is partially performed by a fuzzy operator, which chooses the useful data from all the available data and then it converts it into data objects. Consequently, a list of episodes is created and is used for realisation.

Gkatzia et al.\ \shortcite{Gkatzia2016} present two rule-based approaches to weather forecast generation. The first approach (WMO-based) accounts for the uncertainty present in weather forecasts and it uses the guidelines (as rules) offered by the World Meteorological Organisation  \cite{WMO:2008}for referring to uncertain information. The second approach (NATURAL) imitates the way experts (i.e.\ professional weather forecasters) choose the content to be talked about in the weather forecasts. This approach is more natural than the WMO-based, in that the probabilities are mapped to linguistic interpretation of weather (e.g.\ ``sunny spells") rather than the linguistic mapping of uncertainty (e.g.\ ``likely"). 

Deriving rules by working with experts or by acquiring knowledge from expert generated corpora are two popular practices in developing rule-based data-to-text systems. These approaches offer many benefits as we discuss in Section \ref{litConclusions}.

\section{Trainable Approaches to Content Selection} \label{statistical}

This section discusses trainable approaches to content selection. Trainable approaches to Natural Language Generation often treat content selection and surface realisation in a unified manner. Therefore, here we discuss systems that learn how to choose content, either as an independent task or jointly with surface realisation. Table \ref{trainableTable} summarises trainable approaches to NLG. 

\begin{table*}[ht!]
\centering
\caption{Trainable approaches to content selection in data-to-text systems. In all cases, the data sources are database entries, apart from (Lampouras and Androutsopoulos, 2013) who use ontologies.\label{trainableTable}}
\begin{tabular}{|p{4.4cm}| p{2.7cm}|p{3.5cm} |p{3.1cm}|} 
\hline 
Author(s) & Method & Task & Domain  \\  
\hline \hline
\cite{Mellish1998} & Over-generate and Re-rank & Content Selection & Item descriptions\\ \hline
 \cite{Duboue2002}& Genetic Algorithms & Content Selection & Health \\ \hline
\cite{Duboue2003} & Classification  & Content Selection & Biographical Descriptions \\ \hline
\cite{Barzilay2005} & Classification & Content Selection & Sports  \\ \hline
\cite{Barzilay2004b} & Hidden Markov Models (HMMs) & Content Selection, Ordering, Summarisation & Earthquakes, Clashes, Drugs, Finance, Accidents\\ \hline
\cite{Liang2009} & HMMs & Content Selection & Sportscasting, Weather \\ \hline
\cite{Angeli2010} & HMMs with Log-linear models & Content Selection & Sportscasting, Weather \\ \hline
\cite{Konstas2012}& Structured Perceptron & Content Selection & Flights \\ \hline
\cite{Lampouras2013} & Integer Linear Programming & Content Selection, Lexicalisation and Sentence aggregation & Wine descriptions \\ \hline
\cite{Kondadadi2013} & Support Vector Machines & Content Selection, Realisation & Biography, Weather \\ \hline
\cite{Sowdaboina2014} & Neural Networks & Content Selection & Weather \\ \hline
\cite{Gkatzia2013enlg} & Reinforcement Learning & Content Selection & Student Feedback \\ \hline
\cite{Gkatzia2014acl} & Multi-label Classification & Content Selection & Student Feedback \\ \hline
\cite{Mahapatra2016} &  Multi-partite Graphs & Content Selection & Weather Forecasts \\ \hline
\end{tabular}
\end{table*}

Trainable approaches to NLG have been initially introduced in sentence level \cite{Hatzivassiloglou1995}. Langkilde and Knight \shortcite{Langkilde1998} introduce a trainable approach to Natural Language Generation, which works in two steps. Initially, possible utterances are generated and then they are ranked according to probabilities derived through corpus analysis. Mellish et al.\ \shortcite{Mellish1998} describe a similar stochastic approach based on over-generate and re-rank. The technique is applied in the context of text planning and is used to select the best of the candidate solutions (candidate solutions are generated and then the one to be present in the output is selected stochastically). A similar approach has been later used by Stent et al.\ \shortcite{sparky} for sentence level generation. Similarly, Duboue and McKeown \shortcite{Duboue2002} present an approach to content planning using Genetic Algorithms that is able to identify common patterns in the data.

Duboue and McKeown \shortcite{Duboue2003} present a content selection approach where the available content consists of a corpus of text expressed as semantic features. They treated content selection as a classification task where the objective is to decide whether a database entry should be included in the output or not.

Barzilay and Lapata \shortcite{Barzilay2005} propose a collective content selection approach which is a classification task that makes content decisions in a collective way. Their approach initially considers an ``individual preference score'', which is defined as the preference of the entity to be chosen and it is estimated by: (1) the values of entity features, and (2) the potential association of similar entities. This method has been applied in sports domain where the data can be related in a timely manner, i.e.\ one player's action can cause the injury of another.  The collective content selection approach differs from Duboue and McKeown's \shortcite{Duboue2003} approach in that it allows contextual dependencies because the entries are selected depending on each other and not isolated. 

Barzilay and Lee \shortcite{Barzilay2004b} treats content selection as HMMs, where states correspond to information types and state transitions define the potential ordering. The state transition probabilities define the chance to change from a given topic to another. Liang et al.\ \shortcite{Liang2009} present a model for generation using a 3-tier HMMs in order to address the task of segmenting the utterances, mapping the sentences to meaning representations and choosing the content for generation. The aim of this model is to effectively cope with the segmentation, the grouping of relevant facts, and the alignment of the segmentation results to facts. For this purpose, it is assumed that a world state is represented by records and text, and each record is comprised of fields and their values. For example, in the weather domain, the text is the weather forecast, the records are the different weather attributes such as rain chance, temperature or wind speed, the fields can be the maximum or minimum temperature or wind speed and the values the numerical or categorical values. The parameters of this model are calculated through an EM algorithm and the model is tested in three domains in order to prove its generic nature: Robocup Sportcasting, Weather Reports and NFL Recaps. The process starts with the record selection, e.g. the temperature is selected for generation, then the field selection, e.g. the minimum temperature, and finally the word selection to be generated, e.g. the numerical values of the minimum temperature. The drawback of this model is that it does not treat record, field and word choices in a unified manner so as to capture potential dependencies. Therefore, Angeli et al.\ \shortcite{Angeli2010} extend this model in order to capture the dependencies between records, fields and text. In their model the generation is regarded as a sequence of decisions. 

Konstas and Lapata \shortcite{Konstas2012} present a framework for content selection by discriminatively re-ranking content using the structured perceptron for learning. In this framework, content features are seen as a hypergraph where nodes denote words. Graphical models have been used for NLG in dialogue systems as well, e.g. \cite{Dethlefs2011}. 

Lampouras and Androutsopoulos \shortcite{Lampouras2013} present an Integer Linear Programming model for generation. Their model combines content selection, lexicalisation and sentence aggregation. The ultimate goal of this method is to produce compact text with as short length as possible given an entity of OWL ontology and a set of OWL axioms (facts). 

More recently classifiers have been used for content selection to decide whether an element should be mentioned in the summary or not. Sowdaboina et al\ \shortcite{Sowdaboina2014} use neural networks for content selection in the domain of weather forecasts. Kondadadi et al.\ \shortcite{Kondadadi2013} report a statistical NLG framework for both content planning and realisation. Content is represented as semantic annotations and the realisation is performed using templates. The content selection and realisation decisions are learned from an aligned corpus using Support Vector Machines for modelling the generation and for creating a statistical model. Kondadadi et al.\ \shortcite{Kondadadi2013} do not report using other algorithms for generation. 

Gkatzia et al.\ \shortcite{Gkatzia2014acl} present and compares two trainable approaches to content selection. The first one uses multi-label classification to collectively learn the content to be chosen. In this framework, the learning task is formulated as follows: ``given a set of $X$ time-series factors, select the content that
is most appropriate to be included in a summary''. Labels represent content, whereas each label is represented by a template, and as a result the task can be seen as a classification task. Because the content should be considered simultaneously, multi-label classification is used. In contrast to traditional single-label classification, where the task is to associate a new observation with a label by selecting from a set of labels $L$, in multi-label classification the task is to associate an observation with a set of labels $Y \subseteq L$ \cite{Tsoumakas07}. The second approach uses Reinforcement Learning \cite{Gkatzia2013enlg} for summarisation of time-series data. Content selection is regarded as a Markov Decision problem where the agent aims ``to learn to take the sequence
of actions that leads to optimal content selection''. 

Finally, Mahapatra et al.\ \shortcite{Mahapatra2016}use multi-partite graphs for generation of weather forecasts. In their framework, a partition set for each attribute in the given non-textual dataset is created, and the content is chosen probabilistically from the graph.

Trainable approaches offer many benefits to the developer, they can be easily transferred between domains, they can generalise for unseen instances and can eliminate the need to work with experts. However, big aligned datasets are needed for this task and the quality of the datasets is not always guaranteed. Section \ref{litConclusions} expands on the benefits and drawbacks of trainable approaches and compares and contrasts them with rule-based approaches.

￼
\section{Adaptive Systems} \label{adaptiveSystems}

\begin{table*}[ht!]
\centering
\caption{NLG systems that use User Models.
\label{UMs}}
\begin{tabular}{|p{5cm}| p{8cm}|} % centered columns (4 columns)
\hline %inserts double horizontal lines
Author(s) & Adaptation goal \\  % inserts table 
%heading
\hline \hline% inserts single horizontal line
\cite {Stock2007} & reason of visit, interest and change of background knowledge during the visit\\ \hline
\cite{Demberg2006} & user preferences on flights: price, number of stops, airport location etc. \\ \hline
\cite{Srini2010} &user's inferred prior knowledge \\ \hline
\cite{Han2014} & latent variables \\ \hline
\cite{Walker2007} & users' preferences \\ \hline
\cite{Mairesse2007} & Big Five Personality traits \\ \hline
\cite{Mahamood2011} & stress levels \\ \hline
\cite{Dethlefs2014eacl} & user ratings \\ \hline
\cite{Gkatzia2016fuzz} & unknown users (from other users' ratings) \\ \hline
\end{tabular}
\end{table*}

Adaptation is an area that has been studied in various fields of Computer Science and Human Computer Interaction and consequently, data-to-text systems have been also concerned with adapting the output to particular users. One of the early approaches to adaptive Natural Language Generation is presented by Reiter et al.\ \shortcite{Reiter1999} for the \textit{STOP} system.  This approach uses rules to map questionnaire answers to surface text. As each questionnaire only applies to a specific user, the output is personalised to a user's specific answers.

The predominant way of adaptive data-to-text system is through user modelling. Table \ref{UMs} summarises NLG systems that use User Models (UMs) along with the information included in the UMs. In the context of museum exhibits, Stock et al.\ \shortcite{Stock2007} describe an adaptive multi-modal interactive system, \textit{PEACH}, that is addressed to museum visitors. It consists of:
\begin{itemize}
\item a virtual agent that assists visitors and attracts their attention, 
\item a user-adaptive video display on a mobile device and
\item a user adaptive summary that is generated at the end of the visit in the museum.
\end{itemize}
A predecessor of PEACH is \textit{Ilex} \cite{ODonnell2001} which generates dynamic context in the domain of a virtual museum. PEACH is based on Ilex but it enhances the output with the tailored video. In Ilex, the generation was tailored to user's specific attributes such as reason of visit, interest and change of background knowledge during the visit. The User Model is rule-based. Other innovations of Ilex include the fact that it allows the users to schedule their path and it makes use of the history to present richer summaries, for example by comparing different exhibits that the user has already visited. The content is selected by ranking content and select the most relevant of them.

Demberg and Moore \shortcite{Demberg2006} also suggest a User Model approach to Information Presentation in the context of flight recommendation. In their approach the selected content is influenced by the attributes a user finds important, such as price, number of stops etc. The novelty here is that in order to increase user confidence, the attributes with low value in the user model are briefly summarised, so as to help the user make an informed decision.

NLG systems have also used User Models (UMs) in order to adapt their linguistic output to individual users \cite{Srini2010,Thompson2004,Zukerman2001}. For instance, \cite{Srini2010} propose a system that adapts the generated referring expressions to the user's inferred prior knowledge of the domain. As a user's prior knowledge can change through interactions, they introduce \textit{dynamic user modelling} which allows to update a User Model after interacting with the user.

Han et al.\ \shortcite{Han2014} suggest the use of latent User Models to NLG. In this framework, instead of directly seeking the users' preferences or the users' knowledge through questionnaires, the UMs are inferred through ``hidden'' information derived from sources such as \textit{Google Analytics}. 

Walker et al.\ \shortcite{Walker2007} present an approach that adapts its surface realisations to individual users' preferences, by using an over-generate and re-rank approach. The ranking step of this approach is influenced by the individual's own preferences and therefore the generated realisations are different for each user. Mairesse and Walker \shortcite{Mairesse2007} present a system that recognises the \textit{Big Five} personality traits and use this information for adapting the surface text to a particular user.

NLG systems can employ different versions of a system for each different user group \cite{Gatt2009,Hunter2011,Mahamood2011}. The BT project uses NLG systems in a Neonatal Intensive Care Unit environment to automatically provide reports to different stakeholders. For example, BT-nurse is addressed to nurses working in NICU whereas BT-family is addressed to the parents and relatives of the baby and is able to further adapt to users' stress levels. 

More recently, Dethlefs et al.\ \shortcite{Dethlefs2014eacl} move away from user modelling by exploiting user ratings to infer users' preferences on utterances describing restaurant suggestions.  In their work, users are initially grouped together regarding their similarity in ratings.  Then, the ratings of all users in the same cluster are used in order to predict the ratings of a specific user. Their approach
is efficient in adapting to users' linguistic preferences only after a few ratings are sourced, which means that the output might not be favourable for a first-time user. Therefore, Gkatzia et al.\ \shortcite{Gkatzia2016fuzz} propose a solution for this issue which eliminates the need of initial ratings for first-time users. This approach optimises for all available user groups \cite{Gkatzia2014his} simultaneously using a multi-objective optimisation approach. Assuming that a first-time user will belong to one of the existing user groups and the output is optimised for all groups, the output will also be optimised for each first-time user.

\section{Evaluation Methods} \label{metrics}

Evaluation is very important for data-to-text systems and it can measure several aspects such as task success, effectiveness and similarity to gold standards. Evaluation can be either intrinsic (e.g.\ automatic metrics and user ratings) or extrinsic (e.g.\ evaluation with users in terms of task success), but the extrinsic evaluation is more important, in order to define for what an application is good for and to identify whether an application fulfils its task requirements. Next, we review some forms of evaluation highlighting in what they are successful of evaluating in relevance with the related work mentioned earlier. The evaluation methods are categorised into intrinsic and extrinsic methods, adopting the terminology used by Belz and Hastie \shortcite{Belz2012}. 

\subsection{Intrinsic methods}

\subsubsection{Output Quality Measures}
Automatic metrics are a type of intrinsic evaluation which assess the similarity of the output to a reference model or assess quality criteria \cite{Belz2012}, such as the translation metrics BLEU, NIST, ROUGE, F-measure etc.
\begin{itemize}
\item BLEU (Bilingual Evaluation Understudy) was initially introduced to machine translation in order to evaluate the output quality of machine translated text by comparing it to a human reference translation, so that ``the closer a machine translation is to a professional human translation, the better it is'' \cite{Papineni2002}. Recently, it has been widely used for the evaluation of data-to-text systems to measure the proximity of a machine generated text to a human generated text, e.g.\ \cite{Angeli2010}. 
\item NIST (named after the US National Institute of Standards and Technology) is based on BLEU but it also assesses how informative an n-gram scoring higher for rarer n-gram occurrences \cite{Doddington2002}.
\item ROUGE (Recall-Oriented Understudy for Gisting Evaluation) package was also initially introduced to machine translation and summarisation communities, however it is widely used in NLG as well. It compares the output generated text against a reference text \cite{Lin2004}. ROUGE is a summarisation evaluation package which consists of several automatic metrics: 1) ROUGE-N, which is based on n-grams, 2) ROUGE-L, which is based on Longest Common Sub- sequence, 3) ROUGE-W, which is based on Weighted Longest Common Subsequence and 4) ROUGE-S, which measures the overlap of skip-bigrams between a generated summary and a reference summary. 
\item F-measure is borrowed from statistics and is based on precision and recall \cite{Olson2008}. This measure has been used to evaluate the content selection in data-to-text systems, e.g.\ \cite{Angeli2010,Gkatzia2014acl}.
\end{itemize}

Another way of intrinsic evaluation is human-assessed evaluation, where humans evaluate the generated output in terms of similarity to a reference summary/translation \cite{Belz2012} as described in \cite{BelzKow2010}. In human-aided Machine Translation, post-editing is used to improve the output after machine translation and thus the generated output can be evaluated \cite{Hutchins1992}. This metric has been used for natural language generation too, as for instance in \cite{Sripada2005}.

Automatic metrics are regarded as ``backup'' metrics and they are used with human evaluations. They are not standalone metrics and their results are not always correlated with human evaluations \cite{Belz2006}. The results of a human evaluation are more important, because what really matters is the usability of a system.

\subsubsection{User-like Measures}
User like measures are used to assess the systems' output or a particular module. For this evaluation, users are asked questions such as ``How useful did you find the summary?'' \cite{Belz2012}. This kind of method is used in \cite{Walker2002} and \cite{Foster2007}, where an adaptive system is compared to a non-adaptive. The benefit of this method is that it can performed very quickly and easily in contrast with extrinsic methods which normally require a carefully designed setup.

\subsection{Extrinsic Evaluation}

\subsubsection{User Task Success Measures}
User task success measures measure the effectiveness of the systems' output for the user, such as decision making, comprehension accuracy etc. \cite{Belz2012}. Such an evaluation is used in BabyTalk \cite{Gatt2009}, where the users are shown two outputs and have to make a decision, so as to measure which output is more efficient and helpful in decision making. Gkatzia et al.\ \shortcite{Gkatzia2015enlgdemo} present a game-based setup for evaluation of data-to-text systems, which measures decision making.

\subsubsection{System Purpose Success Measures}
System purpose success measures evaluate a system by measuring ``whether it can fulfil its initial purpose'' \cite{Belz2012}. Such an evaluation is applied to STOP system \cite{Reiter1999} in order to find out whether the purpose of the system was achieved, i.e.\ to define whether users quit smoking.  Although system purpose success measures is extremely important, it is an expensive and time consuming task. In addition, it is vague whether the system is solely successful or whether there are external factors that influence the outcome. If we consider the STOP project, there is uncertainty of whether someone quitted smoking because of the generated letter or due to other circumstances (such as health issues). 

Extrinsic methods are more powerful in indicating whether a data-to-text systems will be successful and whether the users will get ``added value''. Because of the complex nature of these evaluation setups, evaluators should be very careful when designing measure task success and system purpose success experiments, and they should try to restrict the impact of confounding variables. Gkatzia and Mahamood \cite{Gkatzia2015enlgeval} provide an overview of the evaluation practices used in the field over the past decade. 

\section{Conclusions} \label{litConclusions}

This article reported the two main approaches to content selection for data-to-text generation: rule-based approaches and trainable approaches. It introduced content selection in other domains and it reviewed adaptive NLG systems. Finally, it discussed evaluation metrics and their suitability. 

Both rule-based and trainable approaches provide benefits and suffer from limitations as we will discuss in the following paragraphs and as it is depicted in Table \ref{strengthsLimitations}. 
Regarding content selection, rule-based systems based on crafted rules, corpus analysis and expert consultations (Knowledge Acquisition from experts) are more robust and widely used in industry. In addition, the output produced by rule-based systems is more understandable by humans, with no funny elements as it is fully controlled. These systems can also account for outliers as long as rules have been provided to handle extreme examples of data. However, they may not be able to cover all distinct rules as the number of rules increases analogously to the complexity of the domain. In addition, the cost of developing and maintaining such a system is high comparing to systems that use data-driven approaches, as these systems can be scalable by providing more rules. In addition, rules are domain specific and therefore not transferable to other domains.

Statistical methods and Machine Learning approaches, have been widely used and adopted for NLG in spoken dialogue systems as compared to data-to-text systems. With statistical methods (SM), NLG systems have the potential to be more domain independent, automatically optimised and generalised \cite{Angeli2010,Dethlefs2011,Rieser2010}. Content selection rules learnt from data corpora can be more efficient, easily ported in new applications and cheap. Due to their ability to take into account large corpora, their coverage can be extended by using more training examples. However, they do require large amounts of training data. Statistical methods can be more expressive than rule-based systems in many ways, linguistically and adaptively and offer scalability and flexibility. In addition, statistical approaches can be used without taking into account part of speech tagging, syntactic relations and lexical dependencies, because statistical methods could facilitate the learning of the sequence of the words through a corpus, without needing details about the grammar (as in machine translation). However, if not enough training examples are available, those methods can choose content that is not coherent. Also statistical methods do not require the acquisition of knowledge from experts  who can be hard to recruit. %Lemon \shortcite{Lemon2011}  summarises the advantages of learning approaches to NLG over the template-based and rule-based approaches; the adaptive properties in dialogue context, the data-driven process, the action selection policy using precise mathematical models and the generalisation ability to uncertainty in dialogue states. These advantages also hold for data-to-text systems \cite{Angeli2010}. 

\begin{table*}[ht!]
\centering
\caption{Strengths and limitations of data-to-text systems\label{strengthsLimitations}}{
\begin{tabular}{|p{1.8cm}| p{5.3cm} |p{5.3cm}|} % centered columns (4 columns)
\hline %inserts double horizontal lines
Approaches & Strengths & Limitations \\  % inserts table 
%heading
\hline \hline% inserts single horizontal line
Rule-based & - robust in small domains \newline 
- understandable output \newline   
- thoroughly studied \newline  
- suitable for commercial use  & - expensive \newline - not transferable \newline - number of rules increases analogously to the domain complexity\\ \hline
Trainable   & - cheap \newline  - scalable \newline - methods can be reused for new domains \newline - experts are not required & - can produce non-understandable output \newline - require large datasets \newline - depend on quality of data \\ \hline
\end{tabular}}
\end{table*}

%Regarding surface realisation, corpus analysis methods that use statistical and machine learning techniques provide a cost effective method for summarising information from large corpora. According to \cite{Reiter2003}, corpus analysis must be used when there is a large data set that covers normal and non-usual circumstances, the context of the data set is what we would like to produce and the context is consistent. Indeed, if the corpus does not include alternative representations, there would be inconsistencies when, for instance, the representation of numbers can vary, i.e. 5 instead of 05 while 06 instead of 6 \cite{Reiter2003}. \cite{Belz2008} suggests Probabilistic Context-free Representationally Underspecified (pCRU), a framework for language generation which uses probabilistic context-free grammars in the weather forecast domain in order to generate variable outputs. pCRU outperforms the traditional n-gram algorithms that overgenerate and rank, because the probabilistic decision is informed as it proceeds.

\subsection{How to choose an approach for a particular domain and application?}

Explicating researchers' assumptions and claims about why they think an approach might be suitable (or not) for content selection enables us to spherically view the task and therefore recognise suitable and unsuitable approaches, given the task at hand. The following framework is intended to provide a set of core questions to aid NLG researchers and developers in this process:
\begin{itemize}
\item \textbf{Domain:} Is it a large or a small domain?
\item \textbf{Knowledge (either from data or experts)}: If it is a large domain, is an aligned dataset available? If yes, in what format is the data available? If not, can you easily crowd-source data? If it is a small domain, do you have access to expert knowledge?
\item \textbf{Evaluation:} What is the main purpose of the system? Can you design an evaluation which includes potential users?
\end{itemize}

Having a good understanding of the task (domain, available knowledge and evaluation) greatly helps make decisions on which approaches to use and how to evaluate them. This framework is applicable for not only content selection, but also other NLG tasks.

\section{Recommendations for Future Directions} \label{futureWork}
 
There are several open research questions for data-to-text systems, which can be either specific to content selection or they can be researched jointly with surface realisation.

\begin{itemize}
\item{How to transfer approaches between domains or from one language to another?} One potential direction can be the application of transfer learning approaches \cite{transfer}.

\item{How can textual information better be combined with visual information to achieve better task and system purpose success?} There is evidence that multi-modal systems which combine graphs with language are more effective in decision making than systems which use only graphs or only natural language \cite{Gkatzia2016}. However, little is known on which type of data can be better communicated in a multi-modal way or as texts. 

\item{Can we develop approaches which handle uncertain data?} A vast variety of data is uncertain, either because of its nature (e.g.\ stock market data or weather data) or because of its source (e.g.\ data from web). Data-to-text generation will be benefited from research on methods to handle this data and potentially from effectively communicating uncertainty.

\end{itemize}

\bibliography{biblio}

\begin{thebibliography}{}

\bibitem[\protect\citename{Angeli \bgroup et al.\egroup }2010]{Angeli2010}
Gabor Angeli, Percy Liang, and Dan Klein.
\newblock 2010.
\newblock {A simple domain-independent probabilistic approach to generation}.
\newblock In {\em Conference on Empirical Methods in Natural Language
  Processing (EMNLP)}.

\bibitem[\protect\citename{Banaee \bgroup et al.\egroup }2013]{Banaee2013}
Hadi Banaee, Mobyen~Uddin Ahmed, and Amy Loutfi.
\newblock 2013.
\newblock {Towards NLG for Physiological Data Monitoring with Body Area
  Networks}.
\newblock In {\em 14th European Workshop on Natural Language Generation
  (ENLG)}.

\bibitem[\protect\citename{Barzilay and Lapata}2005]{Barzilay2005}
Regina Barzilay and Mirella Lapata.
\newblock 2005.
\newblock {Collective content selection for concept-to-text generation}.
\newblock In {\em Conference on Human Language Technology and Empirical Methods
  in Natural Language Processing (HLT - EMNLP)}.

\bibitem[\protect\citename{Barzilay and Lee}2004]{Barzilay2004b}
Regina Barzilay and Lillian Lee.
\newblock 2004.
\newblock {Catching the drift: Probabilistic content models, with applications
  to generation and summarization}.
\newblock In {\em Human Language Technology Conference of the North American
  Chapter of the Association for Computational Linguistics}.

\bibitem[\protect\citename{Belz and Hastie}2014]{Belz2012}
Anja Belz and Helen Hastie, 2014.
\newblock {\em {Towards Comparative Evaluation and Shared Tasks for NLG in
  Interactive Systems}}.
\newblock Cambridge University Press.

\bibitem[\protect\citename{Belz and Kow}2010]{BelzKow2010}
Anja Belz and Eric Kow.
\newblock 2010.
\newblock {Extracting parallel fragments from comparable corpora for
  data-to-text generation}.
\newblock In {\em 6th International Natural Language Generation Conference
  (INLG)}.

\bibitem[\protect\citename{Belz and Reiter}2006]{Belz2006}
Anja Belz and Ehud Reiter.
\newblock 2006.
\newblock {Comparing Automatic and Human Evaluation of NLG Systems}.
\newblock In {\em 11th Conference of the European Chapter of the Association
  for Computational Linguistics (ACL)}.

\bibitem[\protect\citename{Belz}2008]{Belz2008}
Anja Belz.
\newblock 2008.
\newblock {Automatic generation of weather forecast tests using comprehensive
  probabilistic generation-space models}.
\newblock {\em Natural Language Engineering}, 14(4):431--455.

\bibitem[\protect\citename{Black \bgroup et al.\egroup }2010]{Black2010}
Rolf Black, Joe Reddington, Ehud Reiter, Nava Tintarev, and Annalu Waller.
\newblock 2010.
\newblock {Using NLG and Sensors to Support Personal Narrative for Children
  with Complex Communication Needs}.
\newblock In {\em NAACL HLT 2010 Workshop on Speech and Language Processing for
  Assistive Technologies}.

\bibitem[\protect\citename{Bouayad-Agha \bgroup et al.\egroup
  }2012]{BouayadAgha2012}
Nadjet Bouayad-Agha, Gerard Casamayor, Leo Wanner, and Chris Mellish.
\newblock 2012.
\newblock {Content Selection from Semantic Web Data}.
\newblock In {\em 7th International Natural Language Generation Conference
  (INLG)}.

\bibitem[\protect\citename{Boyd}1998]{Boyd1998}
Sarah Boyd.
\newblock 1998.
\newblock {TREND: A System for Generating Intelligent Descriptions of
  Time-Series Data}.
\newblock In {\em IEEE International Conference on Intelligent Processing
  Systems}.

\bibitem[\protect\citename{Chen and Mooney}2008]{ChenMooney}
David Chen and Raymond Mooney.
\newblock 2008.
\newblock Learning to sportscast: A test of grounded language acquisition.
\newblock In {\em 25th International Conference on Machine Learning (ICML)}.

\bibitem[\protect\citename{Demberg and Moore}2006]{Demberg2006}
Vera Demberg and Johanna Moore.
\newblock 2006.
\newblock {Information Presentation in Spoken Dialogue Systems}.
\newblock In {\em 11th Conference of the European Chapter of the Association
  for Computational Linguistics (EACL)}.

\bibitem[\protect\citename{Demir \bgroup et al.\egroup }2011]{Demir2011}
Seniz Demir, Sandra Carberry, and Kathleen McCoy.
\newblock 2011.
\newblock {Summarizing Information Graphics Textually}.
\newblock {\em Computational Linguistics}, 38(3):527 -- 574.

\bibitem[\protect\citename{Dethlefs and Cuayahuitl}2011]{Dethlefs2011}
Nina Dethlefs and Heriberto Cuayahuitl.
\newblock 2011.
\newblock {Combining hierarchical reinforcement learning and bayesian networks
  for natural language generation in situated dialogue}.
\newblock In {\em 13th European Workshop on Natural Language Generation
  (ENLG)}.

\bibitem[\protect\citename{Dethlefs \bgroup et al.\egroup
  }2014]{Dethlefs2014eacl}
Nina Dethlefs, Heriberto Cuayahuitl, Helen Hastie, Verena Rieser, and Oliver
  Lemon.
\newblock 2014.
\newblock {Cluster-based Prediction of User Ratings for Stylistic Surface
  Realisation}.
\newblock In {\em 14th Conference of the European Chapter of the Association
  for Computational Linguistics (EACL)}.

\bibitem[\protect\citename{DiMarco \bgroup et al.\egroup }2008]{DiMarco2008}
Chryssane DiMarco, Peter Bray, Dominic Covvey, Donald Cowan, Vic DiCiccio, Joan
  Lipa, and Cathy Yang.
\newblock 2008.
\newblock {Authoring and Generation of Individualised Patient Education
  Materials}.
\newblock {\em Information Technology in Healthcare}, 6(1):63-71.

\bibitem[\protect\citename{Doddington}2002]{Doddington2002}
George Doddington.
\newblock 2002.
\newblock {Automatic evaluation of machine translation quality using n-gram
  co-occurrence statistics}.
\newblock In {\em 2nd international conference on Human Language Technology
  Research (HLT)}.

\bibitem[\protect\citename{Duboue and McKeown}2002]{Duboue2002}
Pablo Duboue and Kathleen~R. McKeown.
\newblock 2002.
\newblock {Content Planner Construction via Evolutionary Algorithms and a
  Coprus-based Fitness Function}.
\newblock In {\em 2nd International Natural Language Generation Conference
  (INLG)}.

\bibitem[\protect\citename{Duboue and McKeown}2003]{Duboue2003}
Pable Duboue and K.R. McKeown.
\newblock 2003.
\newblock {Statistical acquisition of Content Selection Rules for Natural
  Language Generation}.
\newblock In {\em Conference on Human Language Technology and Empirical Methods
  in Natural Language Processing (HLT - EMNLP)}.

\bibitem[\protect\citename{Foster and Oberlander}2006]{Foster2006}
Mary~Ellen Foster and Jon Oberlander.
\newblock 2006.
\newblock {Data-driven generation of emphatic facial displays}.
\newblock In {\em 11th Conference of the European Chapter of the Association
  for Computational Linguistics (EACL)}.

\bibitem[\protect\citename{Foster and Oberlander}2007]{Foster2007}
Mary~Ellen Foster and Jon Oberlander.
\newblock 2007.
\newblock {Corpus-based generation of head and eyebrow motion for an embodied
  conversational agent}.
\newblock {\em Language Resources and Evaluation}, 41(3):305 -- 323.

\bibitem[\protect\citename{Gatt \bgroup et al.\egroup }2009]{Gatt2009}
Albert Gatt, Francois Portet, Ehud Reiter, James Hunter, Saad Mahamood, Wendy
  Moncur, and Somayajulu Sripada.
\newblock 2009.
\newblock {From Data to Text in the Neonatal Intensive Care Unit: Using NLG
  Technology for Decision Support and Information Management}.
\newblock {\em AI Communications}, 22: 153-186.

\bibitem[\protect\citename{Gkatzia and Mahamood}2015]{Gkatzia2015enlgeval}
Dimitra Gkatzia and Saad Mahamood.
\newblock 2015.
\newblock A snapshot of nlg evaluation practices 2005 -- 2014.
\newblock In {\em 15th European Workshop on Natural Language Generation
  (ENLG)}.

\bibitem[\protect\citename{Gkatzia \bgroup et al.\egroup
  }2013]{Gkatzia2013enlg}
Dimitra Gkatzia, Helen Hastie, Srinivasan Janarthatanam, and Oliver Lemon.
\newblock 2013.
\newblock {Generating student feedback from time-series data using
  Reinforcement Learning}.
\newblock In {\em In Proceedings of the 14th European Workshop on Natural
  Language Generation (ENLG)}.

\bibitem[\protect\citename{Gkatzia \bgroup et al.\egroup
  }2014a]{Gkatzia2014acl}
Dimitra Gkatzia, Helen Hastie, and Oliver Lemon.
\newblock 2014a.
\newblock {Comparing Multi-label classification with Reinforcement Learning for
  Summarisation of Time-series data}.
\newblock In {\em 52nd Annual Meeting of the Association for Computational
  Linguistics (ACL)}.

\bibitem[\protect\citename{Gkatzia \bgroup et al.\egroup
  }2014b]{Gkatzia2014inlg}
Dimitra Gkatzia, Helen Hastie, and Oliver Lemon.
\newblock 2014b.
\newblock {Multi-adaptive Natural Language Generation using Principal Component
  Regression}.
\newblock In {\em 8th International Natural Language Generation Conference
  (INLG)}.

\bibitem[\protect\citename{Gkatzia \bgroup et al.\egroup
  }2014c]{Gkatzia2014his}
Dimitra Gkatzia, Verena Rieser, Alexander McSporran, Alistair McGowan, Alasdair
  Mort, and Michaela Dewar.
\newblock 2014c.
\newblock {Generating Verbal Descriptions from Medical Sensor Data: A Corpus
  Study on User Preferences}.
\newblock In {\em BCS Health Informatics Scotland (HIS)}.

\bibitem[\protect\citename{Gkatzia \bgroup et al.\egroup
  }2015]{Gkatzia2015enlgdemo}
Dimitra Gkatzia, Amanda Cercas, Verena Rieser, and Oliver Lemon.
\newblock 2015.
\newblock A game-based setup for data collection and task-based evaluation of
  uncertain information presentation.
\newblock In {\em 15th European Workshop on Natural Language Generation
  (ENLG)}.

\bibitem[\protect\citename{Gkatzia \bgroup et al.\egroup }2016a]{Gkatzia2016}
Dimitra Gkatzia, Oliver Lemon, and Verena Rieser.
\newblock 2016a.
\newblock Natural language generation enhances human decision-making with
  uncertain information.
\newblock In {\em 54rth Annual Meeting of the Association for Computational
  Linguistics (ACL)}.

\bibitem[\protect\citename{Gkatzia \bgroup et al.\egroup
  }2016b]{Gkatzia2016fuzz}
Dimitra Gkatzia, Verena Rieser, and Oliver Lemon.
\newblock 2016b.
\newblock How to talk to stranger: Generating medical reports for first-time
  users.
\newblock In {\em In IEEE World Congress on Computational Intelligence (IEEE
  WCCI) -- Proceedings of FUZZ-IEEE}.

\bibitem[\protect\citename{Gkatzia}2015]{GkatziaThesis}
Dimitra Gkatzia.
\newblock 2015.
\newblock {\em Data-driven approaches to content selection for data-to-text
  generation}.
\newblock {Ph.D.} thesis, School of Computer and Mathematical Sciences, Heriot
  Watt University.

\bibitem[\protect\citename{Grice}1975]{Grice1975}
Paul Grice.
\newblock 1975.
\newblock {Logic and conversation}.
\newblock {\em In Syntax and Semantics}, 3.

\bibitem[\protect\citename{Hallett \bgroup et al.\egroup }2006]{Hallett2006}
Catalina Hallett, Richard Power, and Donia Scott.
\newblock 2006.
\newblock {Summarisation and visualisation of e-health data repositories}.
\newblock In {\em UK E-Science All-Hands Meeting}.

\bibitem[\protect\citename{Han \bgroup et al.\egroup }2014]{Han2014}
Xiwu Han, Somayajulu Sripada, Kit~(CJA) Macleod, and Antonio A.~R. Ioris.
\newblock 2014.
\newblock {Latent User Models for Online River Information Tailoring}.
\newblock In {\em 8th International Natural Language Generation Conference
  (INLG)}.

\bibitem[\protect\citename{Hunter \bgroup et al.\egroup }2011]{Hunter2011}
Jim Hunter, Yvonne Freer, Albert Gatt, Yaji Sripada, Cindy Sykes, and Dave
  Westwater.
\newblock 2011.
\newblock {BT-Nurse: Computer Generation of Natural Language Shift Summaries
  from Complex Heterogeneous Medical Data}.
\newblock {\em American Medical Informatics Association}, 18(5):621-624.

\bibitem[\protect\citename{Hutchins and Somers}1992]{Hutchins1992}
W.~John Hutchins and Harold~L. Somers.
\newblock 1992.
\newblock {\em {An introduction to Machine Translation}}.
\newblock Academic Press.

\bibitem[\protect\citename{Janarthanam and Lemon}2010]{Srini2010}
Srinivasan Janarthanam and Oliver Lemon.
\newblock 2010.
\newblock Adaptive referring expression generation in spoken dialogue systems:
  Evaluation with real users.
\newblock In {\em 11th Annual Meeting of the Special Interest Group on
  Discourse and Dialogue (SIGDIAL)}.

\bibitem[\protect\citename{Janarthanam}2011]{SriniThesis}
Srinivasan Janarthanam.
\newblock 2011.
\newblock {\em Learning user modelling strategies for adaptive referring
  expression generation in spoke dialogue systems}.
\newblock PhD thesis, University of Edinburgh.

\bibitem[\protect\citename{Jialin~Pan and Yang}2010]{transfer}
Sinno Jialin~Pan and Qiang Yang.
\newblock 2010.
\newblock A survey on transfer learning.
\newblock {\em IEEE Transactions on Knowledge and Data Engineering},
  22(10):1345 -- 1359.

\bibitem[\protect\citename{Johnson and Lane}2011]{Johnson2011}
Nicholas Johnson and David Lane.
\newblock 2011.
\newblock Narrative monologue as a first step towards advanced mission debrief
  for {AUV} operator situational awareness.
\newblock In {\em 15th International Conference on Advanced Robotics}.

\bibitem[\protect\citename{Knight and
  Hatzivassiloglou}1995]{Hatzivassiloglou1995}
Kevin Knight and Vasilis Hatzivassiloglou.
\newblock 1995.
\newblock {Two-Level, Many-Paths Generation}.
\newblock In {\em Conference of the Association for Computational Linguistics
  (ACL)}.

\bibitem[\protect\citename{Kondadadi \bgroup et al.\egroup
  }2013]{Kondadadi2013}
Ravi Kondadadi, Blake Howald, and Frank Schilder.
\newblock 2013.
\newblock A statistical nlg framework for aggregated planning and realization.
\newblock In {\em 51st Annual Meeting of the Association for Computational
  Linguistics}.

\bibitem[\protect\citename{Konstas and Lapata}2012]{Konstas2012}
Ioannis Konstas and Mirella Lapata.
\newblock 2012.
\newblock Unsupervised concept-to-text generation with hypergraphs.
\newblock In {\em Conference of the North American Chapter of the Association
  for Computational Linguistics (NAACL)}.

\bibitem[\protect\citename{Kootval}2008]{WMO:2008}
Haleh Kootval, editor.
\newblock 2008.
\newblock {\em Guidelines on Communicating Forecast Uncertainty}. World
  Meteorological Organsation.

\bibitem[\protect\citename{Kukich}1983]{Kukich1983}
Karen Kukich.
\newblock 1983.
\newblock Design of a knowledge-based report generator.
\newblock In {\em 21st Annual Meeting of the Association for Computational
  Linguistics (ACL)}.

\bibitem[\protect\citename{Lampouras and Androutsopoulos}2013]{Lampouras2013}
Gerasimos Lampouras and Ion Androutsopoulos.
\newblock 2013.
\newblock Using integer linear programming in concept-to-text generation to
  produce more compact texts.
\newblock In {\em 51st Annual Meeting of the Association for Computational
  Linguistics (ACL)}.

\bibitem[\protect\citename{Langkilde and Knight}1998]{Langkilde1998}
Irene Langkilde and Kevin Knight.
\newblock 1998.
\newblock Generation that exploits coprus-based statistical knowledge.
\newblock In {\em 36th Annual Meeting of the Association for Computational
  Linguistics and 17th International Conference on Computational Linguistics
  (ACL)}.

\bibitem[\protect\citename{Law \bgroup et al.\egroup }2005]{Law2005}
Anna~S. Law, Yvonne Freer, Jim Hunter, Robert~H. Logie, Neil McIntosh, and John
  Quinn.
\newblock 2005.
\newblock A comparison of graphical and textual presentations of time series
  data to support medical decision making in the neonatal intensive care unit.
\newblock {\em Journal of Clinical Monitoring and Computing}, pages 19:
  183--194.

\bibitem[\protect\citename{Liang \bgroup et al.\egroup }2009]{Liang2009}
Percy Liang, Michael~I. Jordan, and Dan Klein.
\newblock 2009.
\newblock Learning semantic correspondences with less supervision.
\newblock In {\em Joint Conference of the 47th Annual Meeting of the
  Association of Computational Linguistics (ACL) and the 4th International
  Joint Conference on Natural Language Processing (IJNLP)}.

\bibitem[\protect\citename{Lin}2004]{Lin2004}
Chin-Yew Lin.
\newblock 2004.
\newblock Rouge: a package for automatic evaluation of summaries.
\newblock In {\em Workshop on Text Summarization (WAS)}.

\bibitem[\protect\citename{Mahamood and Reiter}2011]{Mahamood2011}
Saad Mahamood and Ehud Reiter.
\newblock 2011.
\newblock Generating affective natural language for parents of neonatal
  infants.
\newblock In {\em 13th European Workshop on Natural Language Generation
  (ENLG)}.

\bibitem[\protect\citename{Mahamood \bgroup et al.\egroup }2014]{Mahamood2014}
Saad Mahamood, William Bradshaw, and Ehud Reiter.
\newblock 2014.
\newblock Generating annotated graphs using the nlg pipeline architecture.
\newblock In {\em 8th International Natural Language Generation Conference
  (INLG)}.

\bibitem[\protect\citename{Mahapatra \bgroup et al.\egroup
  }2016]{Mahapatra2016}
Joy Mahapatra, Sudip Kumar~Naskar, and Sivaji Bandyopadhyay.
\newblock 2016.
\newblock Statistical natural language generation from tabular non-textual
  data.
\newblock In {\em International Natural Language Generation Conference (INLG)}.

\bibitem[\protect\citename{Mairesse and Walker}2007]{Mairesse2007}
Francois Mairesse and Marilyn Walker.
\newblock 2007.
\newblock Personality generation for dialogue.
\newblock In {\em 45th Annual Meeting of the Association for Computational
  Linguistics (ACL)}.

\bibitem[\protect\citename{Mann and Thompson}1988]{Mann1988}
William Mann and Sandra Thompson.
\newblock 1988.
\newblock Rhetorical structure theory: Toward a functional theory of text
  generation.
\newblock {\em Text}, 8(3):243 -- 281.

\bibitem[\protect\citename{Mellish \bgroup et al.\egroup }1998]{Mellish1998}
Chris Mellish, Alistair Knott, Jon Oberlander, and Mick O'Donnell.
\newblock 1998.
\newblock Experiments using stochastic search for text planning.
\newblock In {\em International Conference on Natural Language Generation
  (INLG)}.

\bibitem[\protect\citename{O'donnell \bgroup et al.\egroup }2001]{ODonnell2001}
Mick O'donnell, Chris Mellish, Jon Oberlander, and Alistair Knott.
\newblock 2001.
\newblock Ilex: an architecture for a dynamic hypertext generation system.
\newblock {\em Natural Language Engineering}, 7(3):225 -- 250.

\bibitem[\protect\citename{Olson and Delen}2008]{Olson2008}
David Olson and Dursun Delen.
\newblock 2008.
\newblock {\em Advanced Data Mining Techniques}.
\newblock Springer.

\bibitem[\protect\citename{Papineni \bgroup et al.\egroup }2002]{Papineni2002}
Kishore Papineni, Salim Roukos, Todd Ward, and Wei-Jing Zhu.
\newblock 2002.
\newblock Bleu: a method for automatic evaluation of machine translation.
\newblock In {\em 40th Annual Meeting of the Association for Computational
  Linguistics (ACL)}.

\bibitem[\protect\citename{Peddington and Tintarev}2011]{Peddington2011}
Joe Peddington and Nava Tintarev.
\newblock 2011.
\newblock Automatically generating stories from sen- sor data.
\newblock In {\em 6th International Conference on Intelligent user Interfaces
  (IUI)}.

\bibitem[\protect\citename{Petre}1995]{Petre1995}
Marian Petre.
\newblock 1995.
\newblock Why looking isn't always seeing: Readership skills and graphical
  programming.
\newblock {\em Transactions of the ACM}, 38(6):33--44.

\bibitem[\protect\citename{Portet \bgroup et al.\egroup }2007]{Portet2007}
Francois Portet, Ehud Reiter, Albert Gatt, Jim Hunter, Somayajulu Sripada,
  Yvonne Freer, and Cindy Sykes.
\newblock 2007.
\newblock Automatic generation of textual summaries from neonatal intensive
  care data.
\newblock In {\em 11th Conference on Arti cial Intelligence in Medicine
  (AIME)}.

\bibitem[\protect\citename{Ramos-Soto \bgroup et al.\egroup }2015]{Soto2015}
Alejandro Ramos-Soto, Alberto Bugarin, Senen Barro, and Juan Taboada.
\newblock 2015.
\newblock Linguistic descriptions for automatic generation of textual
  short-term weather forecasts on real prediction data.
\newblock {\em IEEE Transactions on Fuzzy Systems}, 23(1):44 -- 57.

\bibitem[\protect\citename{Reiter and Dale}2000]{Reiter2000}
Ehud Reiter and Robert Dale.
\newblock 2000.
\newblock {\em Building Natural Language Generation systems}.
\newblock Cambridge University Press.

\bibitem[\protect\citename{Reiter and Sripada}2002]{Reiter2002}
Ehud Reiter and Somayajulu Sripada.
\newblock 2002.
\newblock {Should Corpora Texts Be Gold Standards for NLG?}
\newblock In {\em 2nd International Natural Language Generation Conference
  (INLG)}.

\bibitem[\protect\citename{Reiter \bgroup et al.\egroup }1999]{Reiter1999}
Ehud Reiter, Roma Robertson, and Liesl Osman.
\newblock 1999.
\newblock Types of knowledge required to personalise smoking cessation letters.
\newblock In {\em Artificial Intelligence in Medicine: Proceedings of the Joint
  European Conference on Artificial Intelligence in Medicine and Medical
  Decision Making}.

\bibitem[\protect\citename{Reiter \bgroup et al.\egroup }2003]{Reiter2003}
Ehud Reiter, Somayajulu~G. Spirada, and Roma Robertson.
\newblock 2003.
\newblock Acquiring correct knowledge for natural language generation.
\newblock {\em Artificial Intelligence Research}, 18:491--516.

\bibitem[\protect\citename{Reiter}2007]{Reiter2007}
Ehud Reiter.
\newblock 2007.
\newblock {An Architecture for Data-to-Text Systems}.
\newblock In {\em 11th European Workshop on Natural Language Generation
  (ENLG)}.

\bibitem[\protect\citename{Rieser \bgroup et al.\egroup }2010]{Rieser2010}
Verena Rieser, Oliver Lemon, and Xingkun Liu.
\newblock 2010.
\newblock Optimising information presentation for spoken dialogue systems.
\newblock In {\em 48th Annual Meeting of the Association for Computational
  Linguistics (ACL)}.

\bibitem[\protect\citename{Schneider \bgroup et al.\egroup
  }2013]{Schneider2013}
Ann Schneider, P-L Vaudry, Alasdair Mort, Chris Mellish, Ehud Reiter, and
  P~Wilson.
\newblock 2013.
\newblock {MIME - NLG in Pre-hospital Care}.
\newblock In {\em 14th European Workshop on Natural Language Generation
  (ENLG)}.

\bibitem[\protect\citename{Sowdaboina \bgroup et al.\egroup
  }2014]{Sowdaboina2014}
Pranay Kumar~Venkata Sowdaboina, Sutanu Chakraborti, and Somayajulu Sripada.
\newblock 2014.
\newblock Learning to summarize time series data.
\newblock {\em Computational Linguistics and Intelligent Text Processing,
  Lecture Notes in Computer Science}, 8403:515 -- 528.

\bibitem[\protect\citename{Sripada and Gao}2007]{Sripada2007}
Somayajulu Sripada and Gao Gao.
\newblock 2007.
\newblock Summarizing dive computer data: A case study in integrating textual
  and graphical presentations of numerical data.
\newblock In {\em Workshop on Multimodal Output Generation (MOG)}.

\bibitem[\protect\citename{Sripada \bgroup et al.\egroup }2001]{Sripada2001}
Somayajulu~G. Sripada, Ehud Reiter, Jim Hunter, and Jin Yu.
\newblock 2001.
\newblock A two-stage model for content determination.
\newblock In {\em 8th European workshop on Natural Language Generation (ENLG)}.

\bibitem[\protect\citename{Sripada \bgroup et al.\egroup }2003]{Sripada2003}
Somayajulu Sripada, Ehud Reiter, Jim Hunter, and Jin Yu.
\newblock 2003.
\newblock Generating english summaries of time series data using the gricean
  maxims.
\newblock In {\em 9th ACM International Conference on Knowledge Discovery and
  Data Mining (KDD)}.

\bibitem[\protect\citename{Sripada \bgroup et al.\egroup }2004]{Sripada2004}
Somayajulu Sripada, Ehud Reiter, I~Davy, and K~Nilssen.
\newblock 2004.
\newblock Lessons from deploying {NLG} technology for marine weather forecast
  text generation.
\newblock In {\em PAIS session of ECAI-2004:760-764}.

\bibitem[\protect\citename{Sripada \bgroup et al.\egroup }2005]{Sripada2005}
Somayajulu~G. Sripada, Ehud Reiter, and Lezan Hawizy.
\newblock 2005.
\newblock {Evaluation of an NLG system using post-edit data}.
\newblock In {\em International Joint Conference on Arti cial Intelligence
  (IJCAI)}.

\bibitem[\protect\citename{Stent \bgroup et al.\egroup }2004]{sparky}
Amanda Stent, Rashmi Prasad, and Marilyn Walker.
\newblock 2004.
\newblock Trainable sentence planning for complex information presentation in
  spoken dialog systems.
\newblock In {\em Proceedings of Association for Computational Linguistics
  (ACL)}.

\bibitem[\protect\citename{Stock \bgroup et al.\egroup }2007]{Stock2007}
Oliviero Stock, Massimo Zancanaro, Paolo Busetta, Charles Callaway, Antonio
  Krger, Michael Kruppa, Tsvi Kuflik, Elena Not, and Cesare Rocchi.
\newblock 2007.
\newblock Adaptive, intelligent presentation of information for the museum
  visitor in peach.
\newblock {\em User Modeling and User-Adapted Interaction}, 17(3):257--304.

\bibitem[\protect\citename{Thomas \bgroup et al.\egroup }2010]{Thomas2010}
Kavita Thomas, Somayajulu Sripada, and Matthijs~L. Noordzij.
\newblock 2010.
\newblock Atlas.txt: Exploring linguistic grounding techniques for
  communicating spatial information to blind users.
\newblock In {\em Universal Access in the Information Society. DOI:
  10.1007/s10209-010-0217-5}.

\bibitem[\protect\citename{Thompson \bgroup et al.\egroup }2004]{Thompson2004}
Cynthia~A. Thompson, Mehmet~H. Goker, and Pat Langley.
\newblock 2004.
\newblock A personalised system for conversational recommendations.
\newblock {\em Journal of Artificial Intelligence Research}, 21(1):393 -- 428.

\bibitem[\protect\citename{Tintarev \bgroup et al.\egroup }2016]{Tintarev2016}
Nava Tintarev, Ehud Reiter, Rolf Black, Annalu Waller, and Joe Reddington.
\newblock 2016.
\newblock {Personal storytelling: Using Natural Language Generation for
  children with complex communication needs, in the wild...}
\newblock {\em International Journal of Human-Computer Studies}, (92-93):1--16.

\bibitem[\protect\citename{Tsoumakas and Katakis}2007]{Tsoumakas07}
Grigorios Tsoumakas and Ioannis Katakis.
\newblock 2007.
\newblock Multi-label classification: An overview.
\newblock {\em Int J Data Warehousing and Mining}, 3(3):1--13.

\bibitem[\protect\citename{Turner \bgroup et al.\egroup }2008]{Turner2008inlg}
Ross Turner, Somayajulu Sripada, Ehud Reiter, and Ian Davy.
\newblock 2008.
\newblock Using spatial reference frames to generate grounded textual summaries
  of georeferenced data.
\newblock In {\em 5th International Natural Language Generation Conference
  (INLG)}.

\bibitem[\protect\citename{van~den Meulen \bgroup et al.\egroup
  }2010]{Meulen2010}
Marian van~den Meulen, Robert Logie, Yvonne Freer, Cindy Sykes, Neil McIntosh,
  and Jim Hunter.
\newblock 2010.
\newblock When a graph is poorer than 100 words: A comparison of computerised
  natural language generation, human generated descriptions and graphical
  displays in neonatal intensive care.
\newblock {\em Applied Cognitive Psychology}, 24:77 -- 89.

\bibitem[\protect\citename{Walker \bgroup et al.\egroup }2002]{Walker2002}
Marilyn~A. Walker, Owen~C. Rambow, and Monica Rogati.
\newblock 2002.
\newblock Training a sentence planner for spoken dialogue using boosting.
\newblock {\em Computer Speech and Language}, 16:409 -- 433.

\bibitem[\protect\citename{Walker \bgroup et al.\egroup }2007]{Walker2007}
Marilyn Walker, Amanda Stent, Francois Mairesse, and Rashmi Prasad.
\newblock 2007.
\newblock Individual and domain adaptation in sentence planning for dialogue.
\newblock {\em Artificial Intelligence Research (JAIR)}, 30:413 -- 456.

\bibitem[\protect\citename{Yu \bgroup et al.\egroup }2007]{Yu2007}
Jin Yu, Ehud Reiter, Jim Hunter, and Chris Mellish.
\newblock 2007.
\newblock Choosing the content of textual summaries of large time-series data
  sets.
\newblock {\em Journal Natural Language Engineering}, 13(1).

\bibitem[\protect\citename{Zukerman and Litman}2001]{Zukerman2001}
Ingrid Zukerman and Diane Litman.
\newblock 2001.
\newblock Natural language processing and user modeling: Synergies and
  limitations.
\newblock {\em In User Modeling and User-Adapted Interaction}, 11(1-2):129 --
  158.

\end{thebibliography}
\bibliographystyle{acl2016}

\appendix

\section{Available Datasets}
\begin{table}[h]
\caption{Available datasets for trainable data-to-text systems\label{trainableDatasets}}{
\begin{tabular}{|p{2.4cm}| p{2cm} |p{7.3cm}|} % centered columns (4 columns)
\hline %inserts double horizontal lines
Author(s) & Domain & URL (if available) \\  % inserts table 
%heading
\hline \hline% inserts single horizontal line
\cite{Belz2008} &  Weather  & \url{http://www.itri.brighton.ac.uk/home/Anja.Belz/Prodigy/}  \\ \hline

\cite{Liang2009} &  Weather  & \url{http://cs.stanford.edu/~pliang/papers/} \\ \hline

\cite{ChenMooney} & Sportscasting & \url{http://www.cs.utexas.edu/~ml/clamp/sportscasting/} \\ \hline
\end{tabular}}
\end{table}

\end{document}